\documentclass[10pt,letterpaper]{article}
\usepackage{pgfplots}
\usepackage{xcolor} % 引入颜色包
\usepackage{algorithm}
\usepackage{algpseudocode}
\usepackage{amsmath} 
\usepackage{booktabs} % For prettier tables
\usepackage{multirow} % Required for multirow cells
\usepackage{amssymb} % For \uparrow symbol
\usepackage{graphicx}
\usepackage{subfig}
\newcommand\blfootnote[1]{%
  \begingroup
  \renewcommand\thefootnote{}\footnote{#1}%
  \addtocounter{footnote}{-1}%
  \endgroup
}
\usepackage{cogsci}
\usepackage{hyperref}
\hypersetup{
    colorlinks=true,
    linkcolor=red,   % 可以根据需要设置链接颜色
    citecolor=black,  % 可以根据需要设置引用链接颜色
    urlcolor=black,    % 可以根据需要设置URL链接颜色
}

\cogscifinalcopy % Uncomment this line for the final submission 

\usepackage{pslatex}
\usepackage{apacite}
\usepackage{float} % Roger Levy added this and changed figure/table
\usepackage{amsmath}

\title{CORE: Mitigating Catastrophic Forgetting in Continual Learning through Cognitive Replay}

\author{{\large Jianshu Zhang$^{\dagger}$} \and {\large Yankai Fu$^{\dagger}$} \and {\large Ziheng Peng$^{\dagger}$} \and  {\large \bf Dongyu Yao} \
and {\large \bf Kun He$^{*}$}\\
        {\large School of Cyber Science and Engineering, Wuhan University, China} }

\begin{document}

\maketitle

\begin{abstract}
\blfootnote{$\dagger$ Equal contribution, *Corresponding author}
This paper introduces a novel perspective to significantly mitigate catastrophic forgetting in continuous learning (CL), which emphasizes models' capacity to preserve existing knowledge and assimilate new information. 
Current replay-based methods treat every task and data sample equally and thus can not fully exploit the potential of the replay buffer.
In response, we propose \textbf{CO}gnitive \textbf{RE}play (CORE), which draws inspiration from human cognitive review processes. 
CORE includes two key strategies: Adaptive Quantity Allocation and Quality-Focused Data Selection. 
The former adaptively modulates the replay buffer allocation for each task based on its forgetting rate, while the latter guarantees the inclusion of representative data that best encapsulates the characteristics of each task within the buffer.
Our approach achieves an average accuracy of 37.95\% on split-CIFAR10, surpassing the best baseline method by 6.52\%. 
Additionally, it significantly enhances the accuracy of the poorest-performing task by 6.30\% compared to the top baseline. 
Code is available at \href{https://github.com/sterzhang/CORE}{https://github.com/sterzhang/CORE}.

\end{abstract}

\section{Introduction}
\begin{figure}
    \centering
    \includegraphics[width=0.87\linewidth]{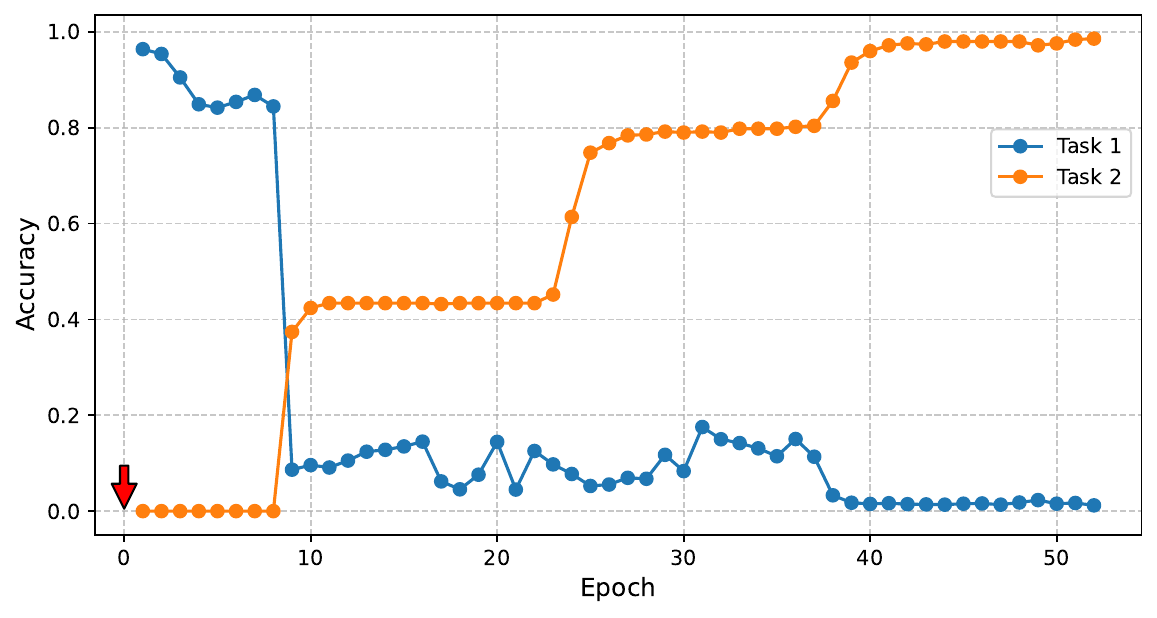}
    \caption{Illustration of Catastrophic Forgetting, which shows Task 1's rapid decline in accuracy when Task 2 starts to train (marked by the red arrow).}
    \label{cata}
\end{figure}

Deep neural networks have shown remarkable capabilities across many tasks~\cite{sze2017efficient}. 
However, in real-world applications, models must continuously adapt to new tasks rather than remain static.
\emph{Continual learning} ~\cite{van2019three} has emerged as a way to dynamically update neural networks, where models continuously acquire knowledge from a stream of tasks~\cite{Chen_Liu_2020}. 
Yet this learning paradigm faces a significant challenge of \emph{Catastrophic Forgetting}~\cite{mccloskey1989catastrophic,french1999catastrophic}, where models tend to lose previous knowledge when learning new tasks~\cite{French_1999}, as shown in Figure \ref{cata}.
This problem highlights the urgent need for effective approaches that can help retain old knowledge when assimilating new information.

Numerous studies propose replay-based methods that attempt to mitigate catastrophic forgetting by replaying old data to review old tasks~\cite{Mnih_Kavukcuoglu_Silver_Graves_Antonoglou_Wierstra_Riedmiller_2013,rolnick2019experience,tiwari2022gcr}.
The core component of these methods is the ``replay buffer'', which contains data from previous tasks.  
However, current methods fail to fully exploit the potential of the replay buffer.
They allocate the replay buffer space indiscriminately across all tasks, failing to differentiate the varying rates of forgetting unique to each task.
In addition, they overlook the quality of the data samples in the buffer, which compromises the effectiveness in data replay.

Inspired by studies that highlight similarities between the human brain and neural networks~\cite{schmidgall2023brain,mcculloch1943logical}, and the remarkable success of applying insights from the human brain to improve the performance of neural networks~\cite{mayo2023multitask,mnih2015human}, we introduce \textbf{CO}gnitive \textbf{RE}play (CORE) to maximize the effectiveness of the replay buffer.
CORE aligns factors of human forgetting with the model's forgetting, specifically cognitive overload and interference. 
It evaluates task-specific accuracy to determine the forgetting rate of previous tasks considering both factors. 
For the newly trained current task, which has not yet experienced cognitive overload, CORE anticipates its potential forgetting by evaluating its interference impact on previous tasks.
We then introduce our innovative Adaptive Quantity Allocation (AQA) strategy, which dynamically allocates buffer space to each task based on the calculated attention derived from their respective forgetting and interference rates.
We also propose the Quality-Focused Data Selection (QFDS) strategy to address the problem of random sample selection in previous methods~\cite{rebuffi2017icarl, rolnick2019experience}.
QFDS ensures that the samples within the replay buffer are more representative and balanced for each task.
These strategies in buffer allocation and data selection enable CORE to outperform traditional baseline methods and establish a more advanced and effective approach to replay-based methods.

Our contributions can be concluded as follows:
\begin{itemize}
    \item We are the first to adopt both minimum and average task accuracy values, along with continual monitoring of each task's performance, which ensures a thorough assessment of model efficacy in continual learning.
    \item  Drawing inspirations
    from human memory mechanisms, CORE implements two innovative strategies, Adaptive Quantity Allocation and Quality Feature-based Data Selection. These strategies collectively refine the utility and efficacy of replay buffers in continual learning models.
    \item Our extensive experiments underscore the effectiveness of CORE. For example, on split-CIFAR10, CORE achieves an average accuracy of 37.95\%, which is about 6.52\% higher than the best baseline method. Remarkably, CORE boosts the accuracy of the most challenging task, exceeding the best baseline by a significant margin of 6.30\%.
\end{itemize}

\section{Theoretical Background}

\subsection{Forgetting in Human Lifelong Learning}

\begin{figure}
    \centering
    \subfloat[Acquisition of continuous tasks in lifelong learning]{
        \includegraphics[width=0.99\linewidth]{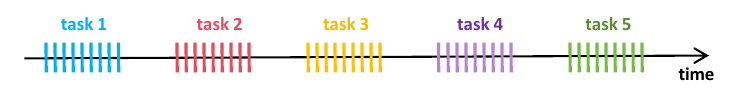}
        \label{Acquisition of continuous tasks in lifelong learning}
    }
    \\
    \subfloat[Cognitive Overload Forgetting]{
        \includegraphics[width=0.4\linewidth]{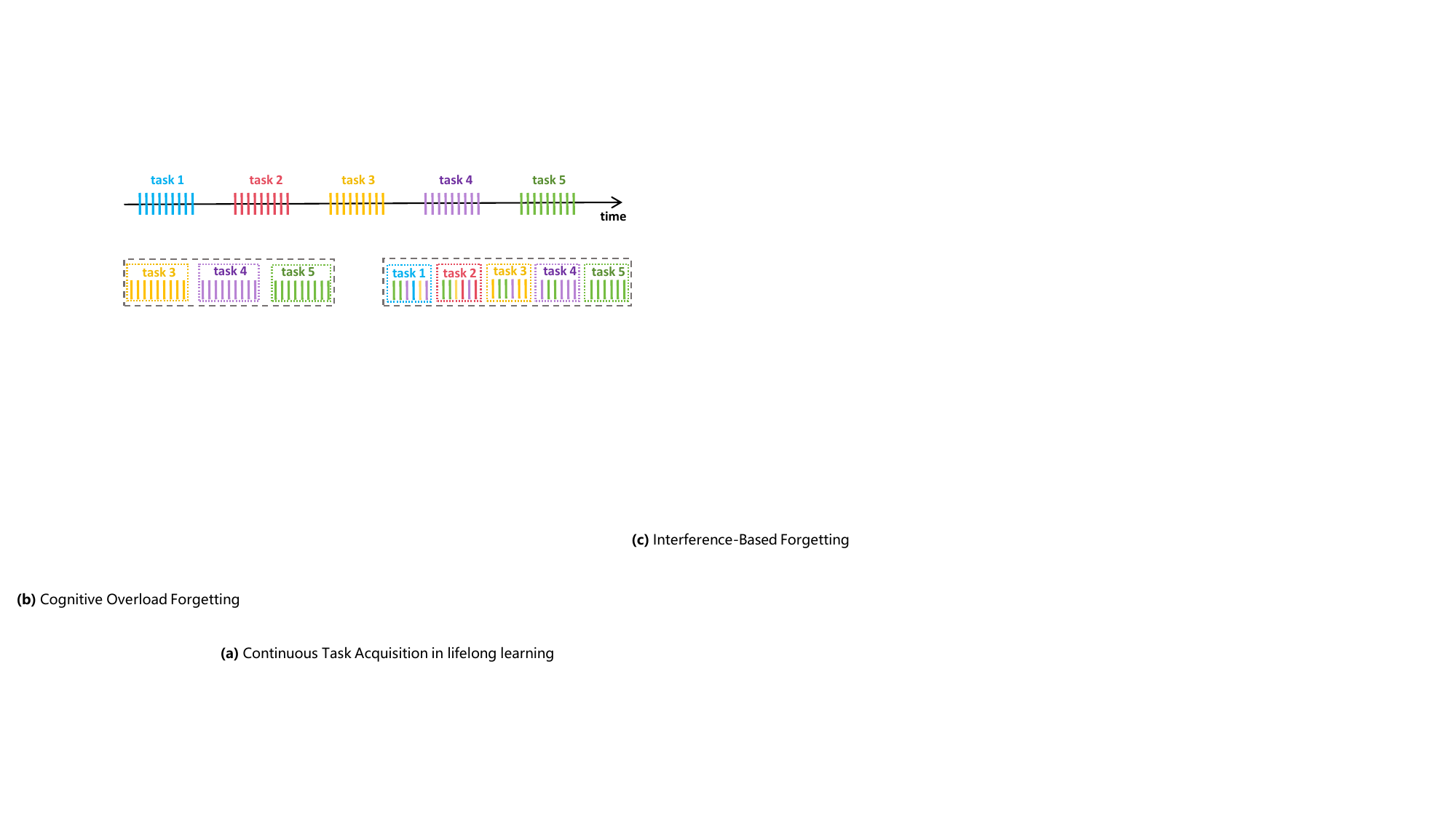}
        \label{Cognitive Overload Forgetting}
    }
    \hspace{5mm}
    \subfloat[Interference-Based Forgetting]{
        \includegraphics[width=0.48\linewidth]{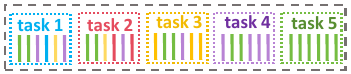}
        \label{Interference-Based Forgetting}
    }
    \caption{Illustration of continuous task acquisition and forgetting mechanisms in lifelong learning.}
    \label{forget}
\end{figure}

In the lifelong learning process of humans (illustrated in Figure \ref{Acquisition of continuous tasks in lifelong learning}), the continual acquisition of new tasks frequently results in the gradual forgetting of previously learned tasks.
According to ~\cite{sadeh2016forgetting}, there are two primary memory types: Recall-based Memory, which involves orthogonal representation and is primarily rooted in the hippocampus~\cite{flesch2022orthogonal}, and Familiarity-based Memory, which is associated with extra-hippocampal structures and employs non-orthogonal representation.

In Recall-based Memory, characterized by distinct and segregated storage, cognitive overload forgetting occurs when new tasks compete for finite capacity of the system. 
This orthogonal representation leads to a scenario where new information directly replaces older memories within the limited available space, as illustrated in Figure \ref{Cognitive Overload Forgetting}.
In Familiarity-based Memory, the vulnerability to interference arises primarily from its non-orthogonal representation, which features overlapping neural pathways. 
This structure allows new information to interfere with established memory. 
Such interference can lead to the deterioration of older memories due to the shared neural connections, as depicted in Figure \ref{Interference-Based Forgetting}.

\subsection{Continual Learning against Forgetting}
In response to the catastrophic forgetting in continual learning, current methods predominantly fall into three categories: \emph{Parameter Isolation}, \emph{Regularization}, and \emph{Data Replay}.

Methods based on parameter isolation, such as Progressive Neural Networks (PNN)~\cite{rusu2016progressive} and context-dependent gating (XdG)~\cite{masse2018alleviating}, segregate parameters for distinct knowledge sets, akin to the orthogonal encoding in Recall-based Memory to minimize task interference. While effective in reducing interference, these approaches face scalability challenges as the knowledge base expands~\cite{wang2023comprehensive}.

Regularization approaches, such as Elastic Weight Consolidation (EWC)~\cite{huszar2018note} and Learning Without Forgetting (LwF)~\cite{li2017learning}, strategically limit important parameter updates to protect previously acquired knowledge.
However, the effectiveness of these methods largely depends on the precise determination of parameter significance in maintaining old knowledge~\cite{de2021continual}.

Methods based on data replay, like Experience Replay (ER)~\cite{rolnick2019experience} and Incremental Classifier and Representation Learning (iCaRL)~\cite{rebuffi2017icarl} utilize historical data to review through a 'replay buffer'. 
The challenge for these approaches lies in effectively utilizing the replay buffer, specifically in how to allocate buffer space and select representative samples strategically.

\subsection{Review with Cognitive Strategies}
In cognitive science, the synergistic application of the \emph{Targeted Recall} and \emph{Spaced Repetition} strategies~\cite{Carpenter_Cepeda_Rohrer_Kang_Pashler_2012}, along with the \emph{Levels of Processing framework}~\cite{Craik_Lockhart_1972}, provides a more effective approach to review.

Targeted Recall emphasizes a more focused and personalized approach that directs heightened attention and resources towards tasks exhibiting higher rates of forgetting.
On the other hand, Spaced Repetition emphasizes the importance of regular, periodic review for sustained memory retention. 
Furthermore, the Levels of Processing framework highlights the role of information quality in learning~\cite{Craik_Lockhart_1972}. 
It distinguishes two types of processing: \emph{Shallow Processing}~\cite{sanford2006shallow}, often resulting in ephemeral memory retention due to its superficial engagement; and \emph{Deep Processing}~\cite{lockhart1990levels}, which encourages an in-depth engagement, leading to the formation of durable and readily retrievable memories.

\begin{figure*}
\begin{center}
\includegraphics[width=0.96\linewidth]{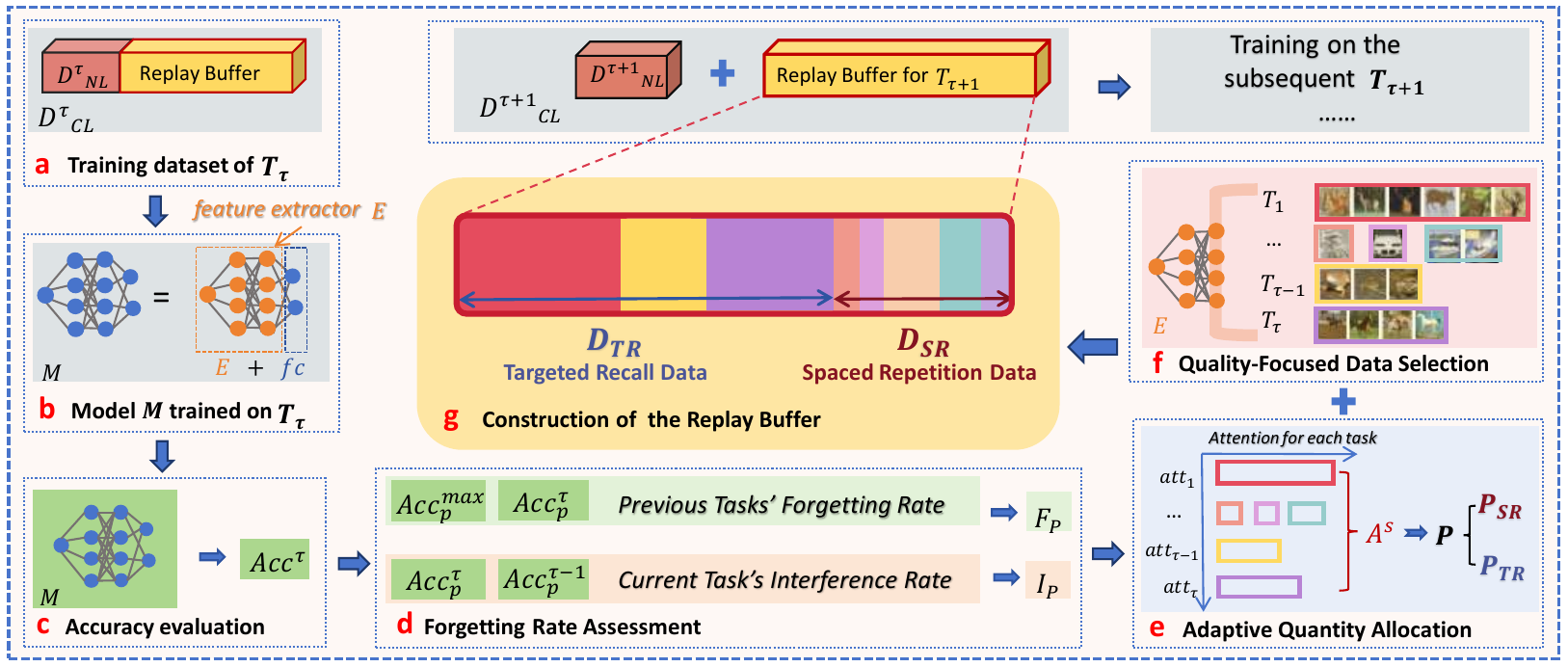}
\end{center}
\caption{Pipeline of CORE. Box.a-g illustrate the process of strategically constructing the replay buffer after the introduction of current task $\mathcal{T}_\tau$ to prepare for the subsequent task $\mathcal{T}_{\tau+1}$.} 
\label{pipeline}
\end{figure*}

\section{Methodology}

\subsection{Overview}

To alleviate catastrophic forgetting in continual learning, we introduce a novel replay-based method named \textbf{CO}gnitive \textbf{RE}play (CORE). 
Figure \ref{pipeline} provides a comprehensive illustration of CORE. 
Initially, the model is trained on $\mathcal{D}_\tau{CL}$, which contains the current task's data and data from the replay buffer. 
After training, an evaluation is conducted to gauge the forgetting rate. 
Utilizing our two principal strategies, we then dynamically allocate buffer space and select high-quality samples for the replay buffer, which is then used for the subsequent task $\mathcal{T}_{\tau+1}$'s training.

\subsection{Cognitive Overload and Interference in Models}
Forgetting in models parallels human cognitive processes, notably cognitive overload and interference. 
Cognitive overload in models occurs when the fixed size of the model, including its architecture and parameter capacity, becomes saturated with new information, hindering its ability to effectively retain and process previously learned information.  
This resembles human cognitive constraints under substantial information loads. 
Interference, inherent in the neural network's interconnected architecture, means even minor parameter adjustments can drastically impact task recall~\cite{Yosinski_Clune_Nguyen_Fuchs_Lipson_2015}. 
This phenomenon reflects interference-based forgetting in human memory, where new learning can obstruct the retention of established knowledge.

\subsection{Forgetting Rate Assessment}
Our methodology strategically constructs the replay buffer, guided by the specific forgetting rates of different tasks. 
To quantify forgetting in previous tasks $\mathcal{P}$, we consider both cognitive overload and interference  (Figure \ref{pipeline} box.d).
For the current task $\mathcal{T}_\tau$, which is newly trained and has not yet exhibited signs of forgetting, we estimate its potential for future forgetting by examining its interference impact on previous tasks.

\subsubsection{Previous Tasks' Forgetting Rate} 
After completing the training of the current task $\mathcal{T}\tau$ (Figure \ref{pipeline} box.b), we perform an accuracy evaluation for each task in the set of previous tasks $\mathcal{P}$. 
All accuracy values during the introduction round of $\mathcal{T}_\tau$ are recorded in the set $Acc^{\tau}$ (Figure \ref{pipeline} box.c).
To compute the forgetting rate $\mathcal{F}_{\mathcal{P}}$ for previous tasks~\cite{wang2023comprehensive}, we compare the historical accuracy of each task with its current performance after training on $\mathcal{T}_\tau$. The forgetting rate $f_p$ for each task is determined as follows:

\begin{equation}
\label{for}
F_{\mathcal{P}} = \left\{ f_{p} = \max_{i \in \{1, \ldots, \tau-1\}} Acc^i_{p} - Acc^{\tau}_{p} \mid p \in \mathcal{P} \right\}
\end{equation}

Here, $f_{p}$ quantifies the extent of performance degradation for each previous task $p$, which effectively reveals the impact of both cognitive overload and interference on forgetting.

\subsubsection{Current Task's Interference Rate} 
Since the cognitive overload of the recently trained task $\mathcal{T}_\tau$ has not yet occurred, we indirectly assess its future forgetting potential by examining how it affects performance on previous tasks$\mathcal{P}$.
This assessment provides insight into how $\mathcal{T}{\tau}$ might be influenced by the data of previous tasks stored in the buffer during the training of $\mathcal{T}{\tau+1}$, which helps us anticipate and address the susceptibility of $\mathcal{T}{\tau}$ to interference-induced forgetting.
The interference rate $\mathcal{I}_{\mathcal{P}}$ is calculated as follows:

\begin{equation}
     I_{\mathcal{P}} = \left\{ i_p = \frac{e^{Acc^{\tau - 1}_{p} - Acc^{\tau}_{p}}}{\sum_{p' \in \mathcal{P}} e^{Acc^{\tau - 1}_{p'} - Acc^{\tau}_{p'}}} \mid  p \in \mathcal{P} \right\}
\end{equation}

Here, \( i_{p} \) quantifies the interference effect on task \( p \) from training \( \mathcal{T}_{\tau} \), which is calculated by the exponential difference in task $p$'s accuracy before (\( Acc^{\tau - 1}_{p} \)) and after (\( Acc^{\tau}_{p} \)) the training of \( \mathcal{T}_{\tau} \). 
This method emphasizes the variation in performance, and the normalization across all tasks in $\mathcal{P}$ ensures a uniform assessment of interference impacts.

\subsection{Adaptive Quantity Allocation}
Adaptive Quantity Allocation (AQA) dynamically manages replay buffer space by calculating each task's attention needs, $\mathcal{A}$, based on its forgetting and interference rates. 
Previous tasks $\mathcal{P}$ are then categorized into two subsets, $\mathcal{P}_{SR}$ for spaced repetition and $\mathcal{P}_{TR}$ for targeted recall, guided by the softmax-transformed attention scores $\mathcal{A}^s$ (Figure \ref{pipeline} box.e).

\subsubsection{Attention Calculation}

For previous tasks, the attention is quantified based on their forgetting rate $\mathcal{F}_\mathcal{P}$ as follows:

\begin{equation}
    \mathcal{A} = \mathcal{A} \cup \biggl\{ att_p = -\log(1-f_p) \mid p \in \mathcal{P}, f_p \in \mathcal{F}_{\mathcal{P}} \biggr\}
\end{equation}

Here, $att_p$ represents the attention allocated to task $p$ according to its forgetting rate $f_p$. 
This ensures that tasks with more significant forgetting get more attention during review.

For the current task $\mathcal{T}_\tau$, the attention is calculated based on its interference with previous tasks:

\begin{equation}
    att_\tau = -log(1-\sum_{i}^{\tau-1} f_p \cdot i_p) , \quad f_p \in \mathcal{F}_\mathcal{P}, i_p \in \mathcal{I}_\mathcal{P}
\end{equation}

In this equation, $att_\tau$ integrates the forgetting rate $\mathcal{F}_\mathcal{P}$ and the interference rate $\mathcal{I}_\mathcal{P}$ over previous tasks $\mathcal{P}$.
This integrated measure assigns appropriate attention to $\mathcal{T}\tau$ to prevent potential forgetting. 
Furthermore, $att_\tau$ is subsequently included in the attention set $\mathcal{A}$ for comprehensive buffer allocation.

\subsubsection{Attention-based Allocation}

After calculating the attention $\mathcal{A}$ , we employ the softmax transformation to normalize $\mathcal{A}$ into a probabilistic distribution $\mathcal{A}^s$:

\begin{equation}
\mathcal{A}^s = \left\{ att^s_i = \frac{e^{{att}_i}}{\sum {e^{{att}_i}}} \mid att_i \in \mathcal{A} \right\}
\end{equation}

Then, we incorporate two key cognitive strategies, Targeted Recall (TR) and Spaced Repetition (SR), to enhance the efficiency of the replay buffer.  
To execute this, we categorize all previous tasks $\mathcal{P}$ \footnote{$\mathcal{T}{\tau}$ is now contained in previous tasks set $\mathcal{P}$.} into two subsets $\mathcal{P}_{SR}, \mathcal{P}_{TR}$.  
This categorization is based on a threshold $ \frac{1}{\lambda\left|\mathcal{P}\right|}$: 
\begin{equation}
p \in \begin{cases}
  \mathcal{P}_{SR}, \quad \text{if } att_p^s \le \frac{1}{\lambda\left|\mathcal{P}\right|} \\
  \mathcal{P}_{TR}, \quad \text{if } att_p^s > \frac{1}{\lambda\left|\mathcal{P}\right|}
\end{cases}
\end{equation}

Here $\left|\mathcal{P}\right|$ represents the number of tasks in the set $\mathcal{P}$. 
$\mathcal{P}_{TR}$, designated for Targeted Recall, contains tasks that require increased attention; while $\mathcal{P}_{SR}$, designated for Spaced Repetition, contains tasks that require minimal attention.

To prevent tasks in $\mathcal{P}_{SR}$ from receiving ineffectively low $att_p^s$, we set a minimum attention threshold:
\begin{equation}
\hat{att_p} = \frac{1}{\lambda \left|\mathcal{P}\right|}, \quad \text{if } p \in \mathcal{P}_{SR}
\end{equation}

After ensuring the minimum allocation of attention to $\mathcal{P}_{SR}$, we focus on proportionally allocating the remaining attention to $\mathcal{P}_{TR}$ based on Targeted Recall (TR):

\begin{equation}
\hat{att_p} = (1 - \frac{|\mathcal{P}_{SR}|}{\lambda\left|\mathcal{P}\right|}) \cdot \frac{att^s_p}{\mathop{\sum}\limits_{p^{'} \in \mathcal{P}_{TR}} att_{p^{'}}^s}, \quad \text{if } p \in \mathcal{P}_{TR}
\end{equation}

In this equation, $\left| \mathcal{P}_{SR} \right|$ represents the number of tasks in $\mathcal{P}_{SR}$. 
This allocation is meticulously designed to effectively implement the Targeted Recall strategy, guaranteeing that tasks more prone to forgetting are allocated with the requisite attention for efficient review and memory reinforcement.

\begin{figure}
    \centering
    \includegraphics[width=0.80\linewidth]{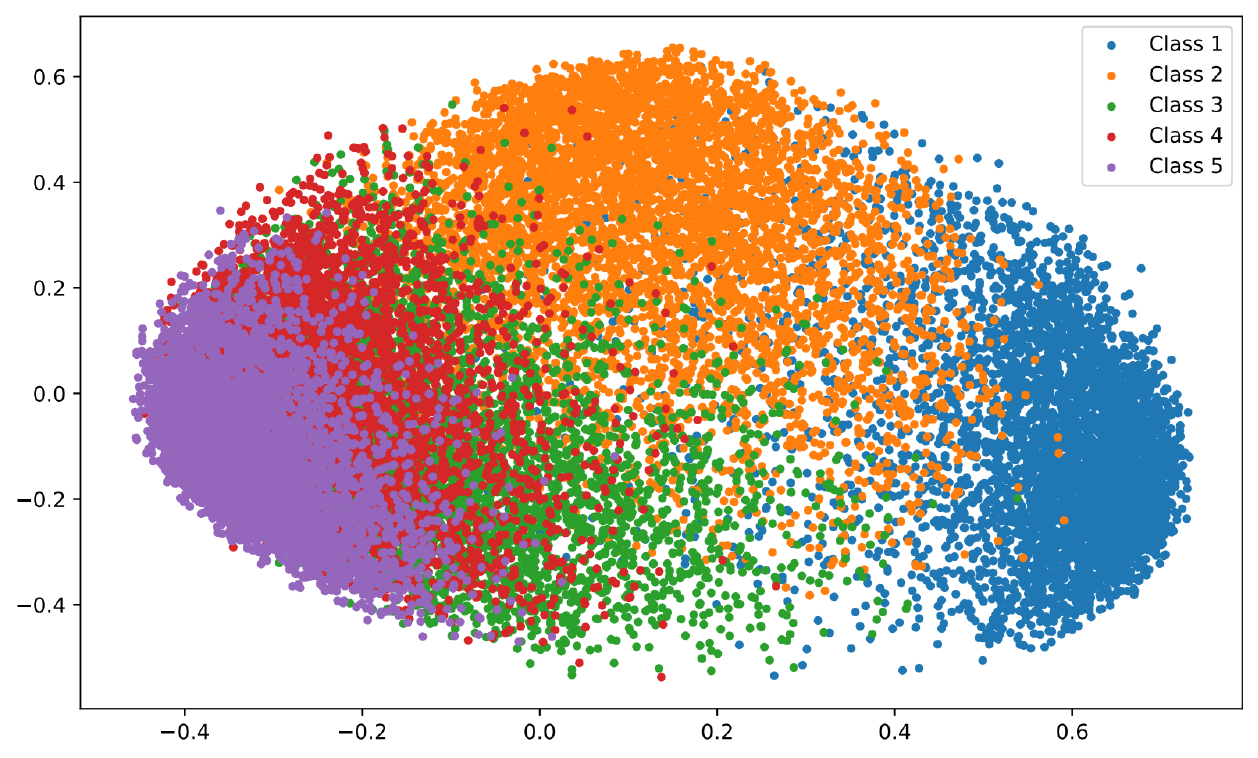}
    \caption{A dimension-reduced feature space for five classes.}
    \label{pca}
\end{figure}

\begin{figure}
    \centering
    \subfloat[Random Data Selection]{
        \includegraphics[width=0.46\linewidth]{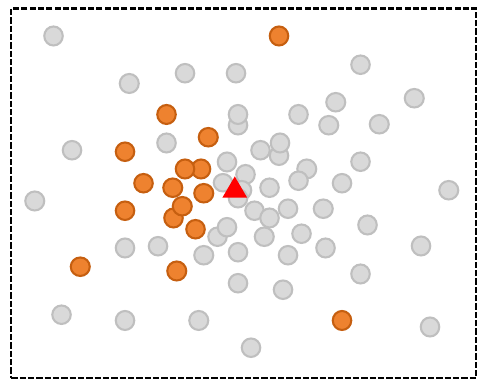}
        \label{Random Data Selection}
    }
    \subfloat[Quality-Focused Data Selection (ours)]{
        \includegraphics[width=0.48\linewidth]{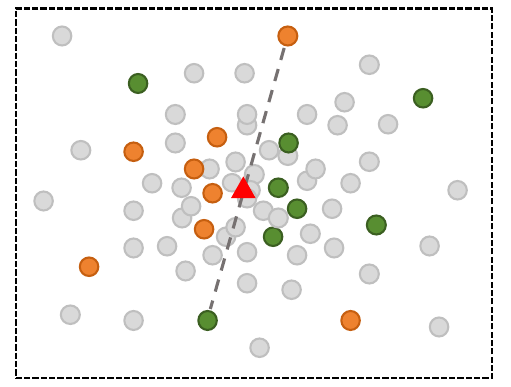}
        \label{Quality-Focused Data Selection}
    }
    \caption{Comparison of data selection methods. The red triangle: the feature center of the class; Orange dots: samples randomly selected; Green dots: samples chosen to balance the sample distribution, which maintain a symmetrical relation (blue dotted line) around the feature center.}
    \label{qfds}
\end{figure}

\subsection{Quality-Focused Data Selection}

In addition to our allocations for $\mathcal{D}_{TR}$ and $\mathcal{D}_{SR}$, we place a strong emphasis on the quality of the data in the replay buffer. 
Inspired by the latent space utilization~\cite{zeiler2014visualizing}, our approach employs a segment of the model as a feature extractor $\mathcal{E}$, which maps data into its latent feature space (Figure \ref{pipeline} box.f). 
The latent feature space (Figure \ref{pca}) aligns with the model's cognitive representation, which enables a more effective data selection.
Additionally, this strategy leverages the model's existing capabilities without additional computational burden. 

We introduce an algorithm to select more representative data for each class by repeatedly performing two steps: 1) randomly selecting a sample from the dataset of the specific class; 2) selecting a subsequent sample that minimizes the distance between the average feature representation of all previously selected samples and the average feature representation of the entire class in the feature space. A detailed procedure is provided in Algorithm \ref{ds}.

Although previous random selection indeed offers equal chances for all samples, it leads to a concentration of samples around certain areas while leaving others sparse (Figure \ref{Random Data Selection}), which may not adequately represent the entire feature distribution.
In contrast, our approach (Figure \ref{Quality-Focused Data Selection}) effectively counters the uneven sample distribution, which guarantees a more uniform coverage of the feature space and a comprehensive representation of the dataset's key features.

\begin{algorithm}
\caption{Feature Based Data Selection}
\label{ds}
\begin{algorithmic}[1]
\Statex \textbf{input} $num$  \hfill//\ Number\ of\ data to be selected
\Statex \textbf{require} $\mathcal{D}$ \quad \quad \quad \quad \quad  //\ Dataset
\Statex \textbf{require} $\mathcal{E} : \mathcal{X} \to \mathcal{R}^{d} $ \quad //\ Feature\ Extractor

\State $\mu \gets \frac{1}{|\mathcal{D}|} \sum\limits_{p \in \mathcal{D}} \mathcal{E} (p)$
\State $SelectData \gets \emptyset$  

\For{$i = 1, 3, 5, \dots, num$} 
    \State $p_i \gets$ RandomlySelect($\mathcal{D}$)
    \State $p_{i + 1} \gets \mathop{\arg\min}\limits_{x \in \mathcal{D}} ||\frac{1}{i + 1} \cdot (\mathop{\sum}\limits_{i = 1}^{i} \mathcal{E} (p_i) + \mathcal{E} (x)) - \mu||$
    \State $SelectData \gets SelectData \cup \{p_i, p_{i + 1}\}$ 
\EndFor

\Statex \textbf{output} $SelectData$ \quad \quad \quad 

\end{algorithmic}
\end{algorithm}

\begin{table*}
\begin{center} 
\caption{The average and minimum accuracy (\%) of $10$ tasks on three datasets. Results are reported as an average of $3$ runs. `$Acc_{avg}$' refers to the average accuracy, and `$Acc_{min}$' refers to the lowest accuracy of tasks. The best results are in bold.}
\label{main_exp}
\vskip 0.12in
\begin{tabular}{@{}c|c|c|c|c|c|c|c|c@{\quad}}
% \begin{tabular}{ccccccccc}
\toprule
Dataset & Metrics & iCaRL & ER  & DGR  & LwF & EWC & Ours & Upper Bound \\
\midrule
\multirow{2}{*}{split-MNIST} 
& $Acc_{avg}$ & 88.00 & 92.39 & 85.46 & 10.00 & 10.00 & \textbf{94.52} & 97.81 \\
& $Acc_{min}$ & 78.97 & 87.47 & 63.85 & 0.00  & 0.00  & \textbf{89.94} & 95.64 \\
\addlinespace 
\midrule
\addlinespace
\multirow{2}{*}{split-CIFAR10} 
& $Acc_{avg}$ & 32.62 & 29.74 & 13.00 & 10.00 & 10.00 & \textbf{37.95} & 65.03 \\
& $Acc_{min}$ &  5.80 & 10.00 & 0.00  & 0.00  & 0.00 & \textbf{16.30} & 44.30 \\
\addlinespace 
\midrule
\addlinespace
\multirow{2}{*}{split-CIFAR100} 
& $Acc_{avg}$ & 24.02 & 25.41 & 6.31 & 10.35 & 6.14 & \textbf{27.24} & 37.95 \\
& $Acc_{min}$ & 15.50 & 18.70 & 0.00 & 0.10  & 0.00 & \textbf{20.40} & 32.60 \\
\bottomrule
\end{tabular}
\end{center} 
\end{table*}

\section{Experiments and Results}

\subsection{Experimental Settings}
We conducted extensive experiments on split-MNIST ~\cite{zenke2017continual}, split-CIFAR-10, and split-CIFAR-100 ~\cite{krizhevsky2009learning} datasets to evaluate our method. 
For a rigorous test of continual learning, we divide each dataset into 10 separate tasks and set the replay buffer size to 500 for all replay-based methods.

To provide a comprehensive evaluation of our approach, we utilize the standard metric of average task accuracy\cite{9349197}.
Moreover, we consider a new metric, the lowest task accuracy across all tasks, which is crucial for identifying  a balanced performance across all tasks, rather than excelling in some while neglecting others.

For a fair comparison with baselines, our experiments are based on the framework provided by \cite{vandeven2022three}, which integrates a range of baseline methodologies in continual learning.
We compare our approach with a range of methodologies in continual learning for a comprehensive assessment. 
This includes replay-based methods such as iCaRL~\cite{rebuffi2017icarl}, ER~\cite{rolnick2019experience}, and DGR~\cite{shin2017continual}, which also utilize a replay buffer like ours. 
To offer a broader perspective on CORE's effectiveness, we compare our approach with baselines based on regularization such as LwF~\cite{li2017learning} and EWC~\cite{huszar2018note}. 
Moreover, we include Joint~\cite{de2021continual} in our comparison to serve as an ideal upper bound that all tasks are trained concurrently.

\subsection{Evaluation Across Baselines}

Table \ref{main_exp} shows the final average accuracy $Acc_{avg}$ of the model, along with the lowest accuracy $Acc_{min}$ across all tasks. 
Across three datasets, we found that regularization methods like LwF and EWC struggle to prevent the forgetting of older tasks, mainly due to the lack of available data. 
This highlights the importance of the replay buffer in knowledge retention. 
Therefore, our analysis focuses mainly on replay-based baselines that employ the replay buffer like ours.

\subsubsection{Experiments on split-MNIST}
We find that though replay-based baselines such as iCaRL and DGR have relatively high average accuracy values, their lowest accuracy values are extremely low. 
The significant difference between DGR's lowest accuracy of 63.85\% and its average accuracy of 85.46\% reveals substantial imbalances in task-specific performance. 
This underscores the importance of the metric $Acc_{min}$ for a holistic evaluation of performance.
In contrast, CORE outperforms the baselines with an average accuracy of 94.52\%, closely approaching the upper-bound of 97.81\%. 
Moreover, the lowest accuracy of all tasks still maintains 89.94\%, largely proving the effectiveness of our approach.

\subsubsection{Experiments on split-CIFAR10}
On a more complex dataset, our approach still achieves an impressive average accuracy of 37.95\%, significantly higher than iCaRL's 32.62\%.
On this dataset, ER's lowest accuracy is only 10.00\%, akin to random guessing. 
Notably, DGR's lowest accuracy plunges to 0.00\%, indicating severe forgetting in the specific task.
Additionally, iCaRL exhibits a staggering disparity of 26.82\% between its average accuracy (32.62\%) and its lowest task accuracy (5.80\%). 
In contrast, CORE maintains a consistent performance, with the lowest task accuracy of 16.30\%, signifying a more balanced approach to knowledge retention.

\subsubsection{Experiments on split-CIFAR100}
It is worth noting that on the larger split-CIFAR100 dataset, where each task contains more classes (i.e. more inference), a more pronounced forgetting occurs. 
This orginal data distribution sort of alleviates the drawbacks of iCaRL and ER's less strategic buffer allocation while undermines the effectiveness of our AQA strategy.
Nevertheless, our proposed QFDS ensures the data samples in the replay buffer are representative, thus guaranteeing CORE's superior performance on split-CIFAR100.

\subsection{Analysis of Progressive Learning Scenarios}

For a comprehensive assessment of the effectiveness of our method, it is insufficient to only consider the final accuracy. 
It is crucial to understand how individual tasks maintain their performance as new tasks are introduced in the process. 
This section aims to show this dynamic by observing the accuracy of the first task $\mathcal{T}_1$ as it trains on nine subsequent tasks.

\begin{figure}
    \centering
    \subfloat[split-MNIST]{
        \includegraphics[width=0.49\linewidth]{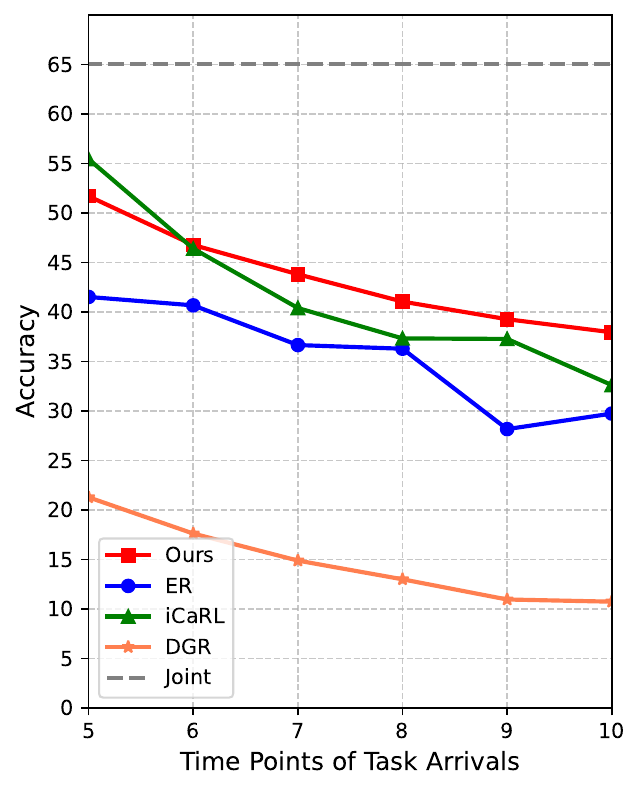}
        \label{split-MNIST}
    }
    \subfloat[split-CIFAR-10]{
        \includegraphics[width=0.49\linewidth]{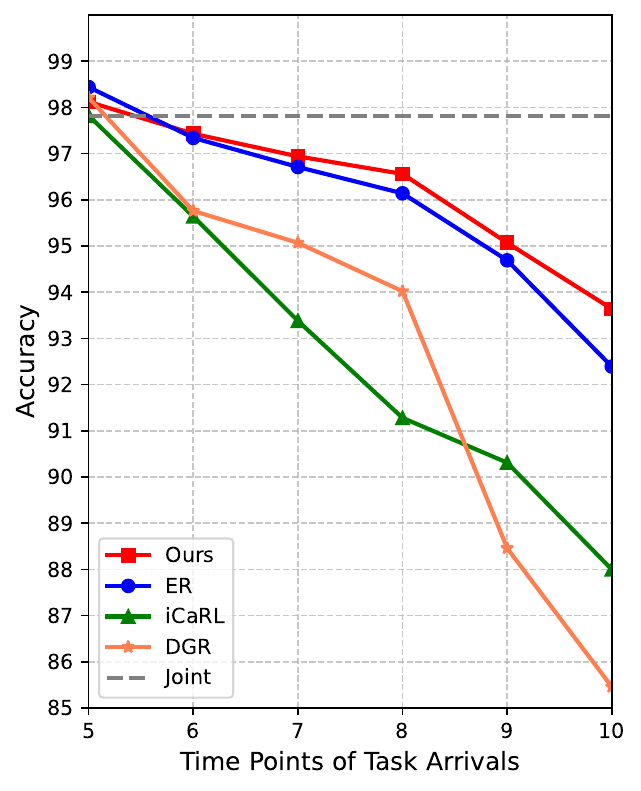}
        \label{split-CIFAR-10}
    }
    \caption{Accuracy trajectories for the first task from the fifth task onward on split-MNIST and split-CIFAR-10.}
    \label{per}
\end{figure}

Figure \ref{per} visualizes the accuracy trajectory of $\mathcal{T}_1$ from the introduction of the fifth task.  
Focusing on this stage is crucial as it marks a point where differences become more apparent.
In the later stages, our method consistently outperforms the baselines, illustrating its robustness in maintaining previous knowledge.
Notably, Figure \ref{split-CIFAR-10} indicates a more stabilized trend in the accuracy decline for our method, suggesting the stability of our approach over time. 
This trend indicates that our approach can effectively manage the influx of multiple tasks without experiencing significant performance drops.

\subsection{Grid Search of Parameter $\lambda$}

In our methodology, the parameter $\lambda$ plays a vital role in setting the minimum attention threshold $\frac{1}{\lambda \left|\mathcal{P}\right|}$, ensuring effective review of each task in the replay buffer. To determine the optimal value of $\lambda$, we conducted a grid search, as outlined in Table \ref{tab:grid_search}. 
The grid search identified $\lambda = 2$ as the optimal value, effectively balancing review efficacy.

\begin{table}
\centering
\caption{Accuracy values for different $\lambda$.}
\label{tab:grid_search}
\vskip 0.12in
\begin{tabular}{cccc}
\hline
\textbf{$\lambda$} & split-MNIST & split-CIFAR10 & split-CIFAR100 \cr \hline
1                 & 92.39                     & 29.74                       &  25.41                      \cr
\textbf{2}                 & \textbf{93.64}                     & \textbf{37.95}                       &  \textbf{27.38}                      \cr
3                 & 93.62                     & 35.55                       &  27.24                      \cr
4                 & 93.04                     & 36.76                       &  26.45                       \cr
5                 & 92.57                     & 34.14                       &  27.19                       \cr \hline
\end{tabular}
\end{table}

\subsection{Ablation Study}
% The ablation study underscores the pivotal roles of CORE's strategies, as removing them results in noticeable performance dips. 
% The absence of Adaptive Quantity Allocation (AQA), which dynamically assigns buffer space based on task forgetfulness, leads to a 5.17\% decrease in average accuracy and a 5.30\% drop in minimum task accuracy. 
% Omitting Quality-Focused Data Selection (QFDS), which ensures the selection of representative data for the buffer, causes a 6.43\% reduction in average accuracy and a 3.90\% decrease in minimum task accuracy. 

% The study reveals that AQA contributes significantly to bolstering the lowest task accuracy, whereas QFDS has a more pronounced effect on lifting the average accuracy across tasks. The most substantial performance decline occurs when both AQA and QFDS are excluded, underscoring their collective criticality in mitigating catastrophic forgetting.
The ablation study reveals CORE's effectiveness, with significant performance reductions when either Adaptive Quantity Allocation (AQA) or Quality-Focused Data Selection (QFDS) is omitted. AQA's absence leads to a 5.17\% decrease in average accuracy and a 5.30\% drop in minimum task accuracy, while omitting QFDS results in a 6.43\% reduction in average accuracy and a 3.90\% decrease in minimum task accuracy. This indicates AQA's crucial role in boosting minimum task accuracy and QFDS's impact on average accuracy, with the greatest performance decline observed when both are excluded, highlighting their combined importance in combating catastrophic forgetting.

\begin{table}[H]
\centering
\caption{Ablation study of CORE on split-CIFAR-10.}
\label{ablation}
\vskip 0.12in
\begin{tabular}{ccc}
\hline
Method               & $Acc_{avg}$               & $Acc_{min}$       \cr \hline
CORE                 & 37.95                     & 16.30                  \cr
CORE w/o AQA         & 32.78                     & 11.00                  \cr
CORE w/o QFDS        & 31.52                     & 12.40                 \cr
CORE w/o QFDS+AQA    & 29.74                     & 10.00                  \cr\hline
\end{tabular}
\end{table}

\section{Conclusion}

In conclusion, CORE mimics human memory mechanism to optimize the utilization of the replay buffer in continual learning, emphasizing both the quantity and quality of data samples within the buffer.
Experiment results confirm our approach's superiority, as it not only achieves the highest average task accuracy but also outperforms on the tasks that are most susceptible to forgetting. 
This advancement establishes CORE as a new benchmark for replay-based paradigms and highlights the benefits of integrating human cognitive strategies into machine learning models.

\section{Acknowledgments}
This research was supported in part by the National Natural Science Foundation of China under grant No. 62172303.

% \newpage
\bibliographystyle{apacite}

\setlength{\bibleftmargin}{.125in}
\setlength{\bibindent}{-\bibleftmargin}

\bibliography{CogSci_Template}

\begin{thebibliography}{}

\bibitem [\protect \citeauthoryear {%
Carpenter%
, Cepeda%
, Rohrer%
, Kang%
\BCBL {}\ \BBA {} Pashler%
}{%
Carpenter%
\ \protect \BOthers {.}}{%
{\protect \APACyear {2012}}%
}]{%
Carpenter_Cepeda_Rohrer_Kang_Pashler_2012}
\APACinsertmetastar {%
Carpenter_Cepeda_Rohrer_Kang_Pashler_2012}%
\begin{APACrefauthors}%
Carpenter, S\BPBI K.%
, Cepeda, N\BPBI J.%
, Rohrer, D.%
, Kang, S\BPBI H\BPBI K.%
\BCBL {}\ \BBA {} Pashler, H.%
\end{APACrefauthors}%
\unskip\
\newblock
\APACrefYearMonthDay{2012}{Sep}{}.
\newblock
{\BBOQ}\APACrefatitle {Using Spacing to Enhance Diverse Forms of Learning: Review of Recent Research and Implications for Instruction} {Using spacing to enhance diverse forms of learning: Review of recent research and implications for instruction}.{\BBCQ}
\newblock
\APACjournalVolNumPages{Educational Psychology Review}{}{}{369–378}.
\newblock
\begin{APACrefDOI} \doi{10.1007/s10648-012-9205-z} \end{APACrefDOI}
\PrintBackRefs{\CurrentBib}

\bibitem [\protect \citeauthoryear {%
Chen%
\ \BBA {} Liu%
}{%
Chen%
\ \BBA {} Liu%
}{%
{\protect \APACyear {2020}}%
}]{%
Chen_Liu_2020}
\APACinsertmetastar {%
Chen_Liu_2020}%
\begin{APACrefauthors}%
Chen, Z.%
\BCBT {}\ \BBA {} Liu, B.%
\end{APACrefauthors}%
\unskip\
\newblock
\APACrefYearMonthDay{2020}{Jan}{}.
\newblock
{\BBOQ}\APACrefatitle {Lifelong Machine Learning} {Lifelong machine learning}.{\BBCQ}
\newblock
\BIn{} \APACrefbtitle {Transfer Learning} {Transfer learning}\ (\BPG~196–208).
\newblock
\begin{APACrefDOI} \doi{10.1017/9781139061773.016} \end{APACrefDOI}
\PrintBackRefs{\CurrentBib}

\bibitem [\protect \citeauthoryear {%
Craik%
\ \BBA {} Lockhart%
}{%
Craik%
\ \BBA {} Lockhart%
}{%
{\protect \APACyear {1972}}%
}]{%
Craik_Lockhart_1972}
\APACinsertmetastar {%
Craik_Lockhart_1972}%
\begin{APACrefauthors}%
Craik, F\BPBI I.%
\BCBT {}\ \BBA {} Lockhart, R\BPBI S.%
\end{APACrefauthors}%
\unskip\
\newblock
\APACrefYearMonthDay{1972}{Dec}{}.
\newblock
{\BBOQ}\APACrefatitle {Levels of processing: A framework for memory research} {Levels of processing: A framework for memory research}.{\BBCQ}
\newblock
\APACjournalVolNumPages{Journal of Verbal Learning and Verbal Behavior}{}{}{671–684}.
\newblock
\begin{APACrefDOI} \doi{10.1016/s0022-5371(72)80001-x} \end{APACrefDOI}
\PrintBackRefs{\CurrentBib}

\bibitem [\protect \citeauthoryear {%
De~Lange%
\ \protect \BOthers {.}}{%
De~Lange%
\ \protect \BOthers {.}}{%
{\protect \APACyear {2021}}%
}]{%
de2021continual}
\APACinsertmetastar {%
de2021continual}%
\begin{APACrefauthors}%
De~Lange, M.%
, Aljundi, R.%
, Masana, M.%
, Parisot, S.%
, Jia, X.%
, Leonardis, A.%
\BDBL {}Tuytelaars, T.%
\end{APACrefauthors}%
\unskip\
\newblock
\APACrefYearMonthDay{2021}{}{}.
\newblock
{\BBOQ}\APACrefatitle {A continual learning survey: Defying forgetting in classification tasks} {A continual learning survey: Defying forgetting in classification tasks}.{\BBCQ}
\newblock
\APACjournalVolNumPages{IEEE transactions on pattern analysis and machine intelligence}{44}{7}{3366--3385}.
\PrintBackRefs{\CurrentBib}

\bibitem [\protect \citeauthoryear {%
De~Lange%
\ \protect \BOthers {.}}{%
De~Lange%
\ \protect \BOthers {.}}{%
{\protect \APACyear {2022}}%
}]{%
9349197}
\APACinsertmetastar {%
9349197}%
\begin{APACrefauthors}%
De~Lange, M.%
, Aljundi, R.%
, Masana, M.%
, Parisot, S.%
, Jia, X.%
, Leonardis, A.%
\BDBL {}Tuytelaars, T.%
\end{APACrefauthors}%
\unskip\
\newblock
\APACrefYearMonthDay{2022}{}{}.
\newblock
{\BBOQ}\APACrefatitle {A Continual Learning Survey: Defying Forgetting in Classification Tasks} {A continual learning survey: Defying forgetting in classification tasks}.{\BBCQ}
\newblock
\APACjournalVolNumPages{IEEE Transactions on Pattern Analysis and Machine Intelligence}{44}{7}{3366-3385}.
\newblock
\begin{APACrefDOI} \doi{10.1109/TPAMI.2021.3057446} \end{APACrefDOI}
\PrintBackRefs{\CurrentBib}

\bibitem [\protect \citeauthoryear {%
Flesch%
, Juechems%
, Dumbalska%
, Saxe%
\BCBL {}\ \BBA {} Summerfield%
}{%
Flesch%
\ \protect \BOthers {.}}{%
{\protect \APACyear {2022}}%
}]{%
flesch2022orthogonal}
\APACinsertmetastar {%
flesch2022orthogonal}%
\begin{APACrefauthors}%
Flesch, T.%
, Juechems, K.%
, Dumbalska, T.%
, Saxe, A.%
\BCBL {}\ \BBA {} Summerfield, C.%
\end{APACrefauthors}%
\unskip\
\newblock
\APACrefYearMonthDay{2022}{}{}.
\newblock
{\BBOQ}\APACrefatitle {Orthogonal representations for robust context-dependent task performance in brains and neural networks} {Orthogonal representations for robust context-dependent task performance in brains and neural networks}.{\BBCQ}
\newblock
\APACjournalVolNumPages{Neuron}{110}{7}{1258--1270}.
\PrintBackRefs{\CurrentBib}

\bibitem [\protect \citeauthoryear {%
R.~French%
}{%
R.~French%
}{%
{\protect \APACyear {1999}}%
}]{%
French_1999}
\APACinsertmetastar {%
French_1999}%
\begin{APACrefauthors}%
French, R.%
\end{APACrefauthors}%
\unskip\
\newblock
\APACrefYearMonthDay{1999}{Apr}{}.
\newblock
{\BBOQ}\APACrefatitle {Catastrophic forgetting in connectionist networks} {Catastrophic forgetting in connectionist networks}.{\BBCQ}
\newblock
\APACjournalVolNumPages{Trends in Cognitive Sciences}{}{}{128–135}.
\newblock
\begin{APACrefDOI} \doi{10.1016/s1364-6613(99)01294-2} \end{APACrefDOI}
\PrintBackRefs{\CurrentBib}

\bibitem [\protect \citeauthoryear {%
R\BPBI M.~French%
}{%
R\BPBI M.~French%
}{%
{\protect \APACyear {1999}}%
}]{%
french1999catastrophic}
\APACinsertmetastar {%
french1999catastrophic}%
\begin{APACrefauthors}%
French, R\BPBI M.%
\end{APACrefauthors}%
\unskip\
\newblock
\APACrefYearMonthDay{1999}{}{}.
\newblock
{\BBOQ}\APACrefatitle {Catastrophic forgetting in connectionist networks} {Catastrophic forgetting in connectionist networks}.{\BBCQ}
\newblock
\APACjournalVolNumPages{Trends in cognitive sciences}{3}{4}{128--135}.
\PrintBackRefs{\CurrentBib}

\bibitem [\protect \citeauthoryear {%
Husz{\'a}r%
}{%
Husz{\'a}r%
}{%
{\protect \APACyear {2018}}%
}]{%
huszar2018note}
\APACinsertmetastar {%
huszar2018note}%
\begin{APACrefauthors}%
Husz{\'a}r, F.%
\end{APACrefauthors}%
\unskip\
\newblock
\APACrefYearMonthDay{2018}{}{}.
\newblock
{\BBOQ}\APACrefatitle {Note on the quadratic penalties in elastic weight consolidation} {Note on the quadratic penalties in elastic weight consolidation}.{\BBCQ}
\newblock
\APACjournalVolNumPages{Proceedings of the National Academy of Sciences}{115}{11}{E2496--E2497}.
\PrintBackRefs{\CurrentBib}

\bibitem [\protect \citeauthoryear {%
Krizhevsky%
, Hinton%
\BCBL {}\ \protect \BOthers {.}}{%
Krizhevsky%
\ \protect \BOthers {.}}{%
{\protect \APACyear {2009}}%
}]{%
krizhevsky2009learning}
\APACinsertmetastar {%
krizhevsky2009learning}%
\begin{APACrefauthors}%
Krizhevsky, A.%
, Hinton, G.%
\BCBL {}\ \BOthersPeriod {.}\end{APACrefauthors}%
\unskip\
\newblock
\APACrefYearMonthDay{2009}{}{}.
\newblock
{\BBOQ}\APACrefatitle {Learning multiple layers of features from tiny images} {Learning multiple layers of features from tiny images}.{\BBCQ}
\newblock

\PrintBackRefs{\CurrentBib}

\bibitem [\protect \citeauthoryear {%
Li%
\ \BBA {} Hoiem%
}{%
Li%
\ \BBA {} Hoiem%
}{%
{\protect \APACyear {2017}}%
}]{%
li2017learning}
\APACinsertmetastar {%
li2017learning}%
\begin{APACrefauthors}%
Li, Z.%
\BCBT {}\ \BBA {} Hoiem, D.%
\end{APACrefauthors}%
\unskip\
\newblock
\APACrefYearMonthDay{2017}{}{}.
\newblock
{\BBOQ}\APACrefatitle {Learning without forgetting} {Learning without forgetting}.{\BBCQ}
\newblock
\APACjournalVolNumPages{IEEE transactions on pattern analysis and machine intelligence}{40}{12}{2935--2947}.
\PrintBackRefs{\CurrentBib}

\bibitem [\protect \citeauthoryear {%
Lockhart%
\ \BBA {} Craik%
}{%
Lockhart%
\ \BBA {} Craik%
}{%
{\protect \APACyear {1990}}%
}]{%
lockhart1990levels}
\APACinsertmetastar {%
lockhart1990levels}%
\begin{APACrefauthors}%
Lockhart, R\BPBI S.%
\BCBT {}\ \BBA {} Craik, F\BPBI I.%
\end{APACrefauthors}%
\unskip\
\newblock
\APACrefYearMonthDay{1990}{}{}.
\newblock
{\BBOQ}\APACrefatitle {Levels of processing: A retrospective commentary on a framework for memory research.} {Levels of processing: A retrospective commentary on a framework for memory research.}{\BBCQ}
\newblock
\APACjournalVolNumPages{Canadian Journal of Psychology/Revue canadienne de psychologie}{44}{1}{87}.
\PrintBackRefs{\CurrentBib}

\bibitem [\protect \citeauthoryear {%
Masse%
, Grant%
\BCBL {}\ \BBA {} Freedman%
}{%
Masse%
\ \protect \BOthers {.}}{%
{\protect \APACyear {2018}}%
}]{%
masse2018alleviating}
\APACinsertmetastar {%
masse2018alleviating}%
\begin{APACrefauthors}%
Masse, N\BPBI Y.%
, Grant, G\BPBI D.%
\BCBL {}\ \BBA {} Freedman, D\BPBI J.%
\end{APACrefauthors}%
\unskip\
\newblock
\APACrefYearMonthDay{2018}{}{}.
\newblock
{\BBOQ}\APACrefatitle {Alleviating catastrophic forgetting using context-dependent gating and synaptic stabilization} {Alleviating catastrophic forgetting using context-dependent gating and synaptic stabilization}.{\BBCQ}
\newblock
\APACjournalVolNumPages{Proceedings of the National Academy of Sciences}{115}{44}{E10467--E10475}.
\PrintBackRefs{\CurrentBib}

\bibitem [\protect \citeauthoryear {%
Mayo%
\ \protect \BOthers {.}}{%
Mayo%
\ \protect \BOthers {.}}{%
{\protect \APACyear {2023}}%
}]{%
mayo2023multitask}
\APACinsertmetastar {%
mayo2023multitask}%
\begin{APACrefauthors}%
Mayo, D.%
, Scott, T\BPBI R.%
, Ren, M.%
, Elsayed, G.%
, Hermann, K.%
, Jones, M.%
\BCBL {}\ \BBA {} Mozer, M.%
\end{APACrefauthors}%
\unskip\
\newblock
\APACrefYearMonthDay{2023}{}{}.
\newblock
{\BBOQ}\APACrefatitle {Multitask learning via interleaving: A neural network investigation} {Multitask learning via interleaving: A neural network investigation}.{\BBCQ}
\newblock
\BIn{} \APACrefbtitle {Proceedings of the Annual Meeting of the Cognitive Science Society} {Proceedings of the annual meeting of the cognitive science society}\ (\BVOL~45).
\PrintBackRefs{\CurrentBib}

\bibitem [\protect \citeauthoryear {%
McCloskey%
\ \BBA {} Cohen%
}{%
McCloskey%
\ \BBA {} Cohen%
}{%
{\protect \APACyear {1989}}%
}]{%
mccloskey1989catastrophic}
\APACinsertmetastar {%
mccloskey1989catastrophic}%
\begin{APACrefauthors}%
McCloskey, M.%
\BCBT {}\ \BBA {} Cohen, N\BPBI J.%
\end{APACrefauthors}%
\unskip\
\newblock
\APACrefYearMonthDay{1989}{}{}.
\newblock
{\BBOQ}\APACrefatitle {Catastrophic interference in connectionist networks: The sequential learning problem} {Catastrophic interference in connectionist networks: The sequential learning problem}.{\BBCQ}
\newblock
\BIn{} \APACrefbtitle {Psychology of learning and motivation} {Psychology of learning and motivation}\ (\BVOL~24, \BPGS\ 109--165).
\newblock
\APACaddressPublisher{}{Elsevier}.
\PrintBackRefs{\CurrentBib}

\bibitem [\protect \citeauthoryear {%
McCulloch%
\ \BBA {} Pitts%
}{%
McCulloch%
\ \BBA {} Pitts%
}{%
{\protect \APACyear {1943}}%
}]{%
mcculloch1943logical}
\APACinsertmetastar {%
mcculloch1943logical}%
\begin{APACrefauthors}%
McCulloch, W\BPBI S.%
\BCBT {}\ \BBA {} Pitts, W.%
\end{APACrefauthors}%
\unskip\
\newblock
\APACrefYearMonthDay{1943}{}{}.
\newblock
{\BBOQ}\APACrefatitle {A logical calculus of the ideas immanent in nervous activity} {A logical calculus of the ideas immanent in nervous activity}.{\BBCQ}
\newblock
\APACjournalVolNumPages{The bulletin of mathematical biophysics}{5}{}{115--133}.
\PrintBackRefs{\CurrentBib}

\bibitem [\protect \citeauthoryear {%
Mnih%
\ \protect \BOthers {.}}{%
Mnih%
\ \protect \BOthers {.}}{%
{\protect \APACyear {2013}}%
}]{%
Mnih_Kavukcuoglu_Silver_Graves_Antonoglou_Wierstra_Riedmiller_2013}
\APACinsertmetastar {%
Mnih_Kavukcuoglu_Silver_Graves_Antonoglou_Wierstra_Riedmiller_2013}%
\begin{APACrefauthors}%
Mnih, V.%
, Kavukcuoglu, K.%
, Silver, D.%
, Graves, A.%
, Antonoglou, I.%
, Wierstra, D.%
\BCBL {}\ \BBA {} Riedmiller, M.%
\end{APACrefauthors}%
\unskip\
\newblock
\APACrefYearMonthDay{2013}{Dec}{}.
\newblock
{\BBOQ}\APACrefatitle {Playing Atari with Deep Reinforcement Learning} {Playing atari with deep reinforcement learning}.{\BBCQ}
\newblock
\APACjournalVolNumPages{arXiv: Learning,arXiv: Learning}{}{}{}.
\PrintBackRefs{\CurrentBib}

\bibitem [\protect \citeauthoryear {%
Mnih%
\ \protect \BOthers {.}}{%
Mnih%
\ \protect \BOthers {.}}{%
{\protect \APACyear {2015}}%
}]{%
mnih2015human}
\APACinsertmetastar {%
mnih2015human}%
\begin{APACrefauthors}%
Mnih, V.%
, Kavukcuoglu, K.%
, Silver, D.%
, Rusu, A\BPBI A.%
, Veness, J.%
, Bellemare, M\BPBI G.%
\BDBL {}others%
\end{APACrefauthors}%
\unskip\
\newblock
\APACrefYearMonthDay{2015}{}{}.
\newblock
{\BBOQ}\APACrefatitle {Human-level control through deep reinforcement learning} {Human-level control through deep reinforcement learning}.{\BBCQ}
\newblock
\APACjournalVolNumPages{nature}{518}{7540}{529--533}.
\PrintBackRefs{\CurrentBib}

\bibitem [\protect \citeauthoryear {%
Rebuffi%
, Kolesnikov%
, Sperl%
\BCBL {}\ \BBA {} Lampert%
}{%
Rebuffi%
\ \protect \BOthers {.}}{%
{\protect \APACyear {2017}}%
}]{%
rebuffi2017icarl}
\APACinsertmetastar {%
rebuffi2017icarl}%
\begin{APACrefauthors}%
Rebuffi, S\BHBI A.%
, Kolesnikov, A.%
, Sperl, G.%
\BCBL {}\ \BBA {} Lampert, C\BPBI H.%
\end{APACrefauthors}%
\unskip\
\newblock
\APACrefYearMonthDay{2017}{}{}.
\newblock
{\BBOQ}\APACrefatitle {icarl: Incremental classifier and representation learning} {icarl: Incremental classifier and representation learning}.{\BBCQ}
\newblock
\BIn{} \APACrefbtitle {Proceedings of the IEEE conference on Computer Vision and Pattern Recognition} {Proceedings of the ieee conference on computer vision and pattern recognition}\ (\BPGS\ 2001--2010).
\PrintBackRefs{\CurrentBib}

\bibitem [\protect \citeauthoryear {%
Rolnick%
, Ahuja%
, Schwarz%
, Lillicrap%
\BCBL {}\ \BBA {} Wayne%
}{%
Rolnick%
\ \protect \BOthers {.}}{%
{\protect \APACyear {2019}}%
}]{%
rolnick2019experience}
\APACinsertmetastar {%
rolnick2019experience}%
\begin{APACrefauthors}%
Rolnick, D.%
, Ahuja, A.%
, Schwarz, J.%
, Lillicrap, T.%
\BCBL {}\ \BBA {} Wayne, G.%
\end{APACrefauthors}%
\unskip\
\newblock
\APACrefYearMonthDay{2019}{}{}.
\newblock
{\BBOQ}\APACrefatitle {Experience replay for continual learning} {Experience replay for continual learning}.{\BBCQ}
\newblock
\APACjournalVolNumPages{Advances in Neural Information Processing Systems}{32}{}{}.
\PrintBackRefs{\CurrentBib}

\bibitem [\protect \citeauthoryear {%
Rusu%
\ \protect \BOthers {.}}{%
Rusu%
\ \protect \BOthers {.}}{%
{\protect \APACyear {2016}}%
}]{%
rusu2016progressive}
\APACinsertmetastar {%
rusu2016progressive}%
\begin{APACrefauthors}%
Rusu, A\BPBI A.%
, Rabinowitz, N\BPBI C.%
, Desjardins, G.%
, Soyer, H.%
, Kirkpatrick, J.%
, Kavukcuoglu, K.%
\BDBL {}Hadsell, R.%
\end{APACrefauthors}%
\unskip\
\newblock
\APACrefYearMonthDay{2016}{}{}.
\newblock
{\BBOQ}\APACrefatitle {Progressive neural networks} {Progressive neural networks}.{\BBCQ}
\newblock
\APACjournalVolNumPages{arXiv preprint arXiv:1606.04671}{}{}{}.
\PrintBackRefs{\CurrentBib}

\bibitem [\protect \citeauthoryear {%
Sadeh%
, Ozubko%
, Winocur%
\BCBL {}\ \BBA {} Moscovitch%
}{%
Sadeh%
\ \protect \BOthers {.}}{%
{\protect \APACyear {2016}}%
}]{%
sadeh2016forgetting}
\APACinsertmetastar {%
sadeh2016forgetting}%
\begin{APACrefauthors}%
Sadeh, T.%
, Ozubko, J\BPBI D.%
, Winocur, G.%
\BCBL {}\ \BBA {} Moscovitch, M.%
\end{APACrefauthors}%
\unskip\
\newblock
\APACrefYearMonthDay{2016}{}{}.
\newblock
{\BBOQ}\APACrefatitle {Forgetting patterns differentiate between two forms of memory representation} {Forgetting patterns differentiate between two forms of memory representation}.{\BBCQ}
\newblock
\APACjournalVolNumPages{Psychological science}{27}{6}{810--820}.
\PrintBackRefs{\CurrentBib}

\bibitem [\protect \citeauthoryear {%
Sanford%
\ \BBA {} Graesser%
}{%
Sanford%
\ \BBA {} Graesser%
}{%
{\protect \APACyear {2006}}%
}]{%
sanford2006shallow}
\APACinsertmetastar {%
sanford2006shallow}%
\begin{APACrefauthors}%
Sanford, A\BPBI J.%
\BCBT {}\ \BBA {} Graesser, A\BPBI C.%
\end{APACrefauthors}%
\unskip\
\newblock
\APACrefYearMonthDay{2006}{}{}.
\newblock
{\BBOQ}\APACrefatitle {Shallow processing and underspecification} {Shallow processing and underspecification}.{\BBCQ}
\newblock
\APACjournalVolNumPages{Discourse Processes}{42}{2}{99--108}.
\PrintBackRefs{\CurrentBib}

\bibitem [\protect \citeauthoryear {%
Schmidgall%
\ \protect \BOthers {.}}{%
Schmidgall%
\ \protect \BOthers {.}}{%
{\protect \APACyear {2023}}%
}]{%
schmidgall2023brain}
\APACinsertmetastar {%
schmidgall2023brain}%
\begin{APACrefauthors}%
Schmidgall, S.%
, Achterberg, J.%
, Miconi, T.%
, Kirsch, L.%
, Ziaei, R.%
, Hajiseyedrazi, S.%
\BCBL {}\ \BBA {} Eshraghian, J.%
\end{APACrefauthors}%
\unskip\
\newblock
\APACrefYearMonthDay{2023}{}{}.
\newblock
{\BBOQ}\APACrefatitle {Brain-inspired learning in artificial neural networks: a review} {Brain-inspired learning in artificial neural networks: a review}.{\BBCQ}
\newblock
\APACjournalVolNumPages{arXiv preprint arXiv:2305.11252}{}{}{}.
\PrintBackRefs{\CurrentBib}

\bibitem [\protect \citeauthoryear {%
Shin%
, Lee%
, Kim%
\BCBL {}\ \BBA {} Kim%
}{%
Shin%
\ \protect \BOthers {.}}{%
{\protect \APACyear {2017}}%
}]{%
shin2017continual}
\APACinsertmetastar {%
shin2017continual}%
\begin{APACrefauthors}%
Shin, H.%
, Lee, J\BPBI K.%
, Kim, J.%
\BCBL {}\ \BBA {} Kim, J.%
\end{APACrefauthors}%
\unskip\
\newblock
\APACrefYearMonthDay{2017}{}{}.
\newblock
{\BBOQ}\APACrefatitle {Continual learning with deep generative replay} {Continual learning with deep generative replay}.{\BBCQ}
\newblock
\APACjournalVolNumPages{Advances in neural information processing systems}{30}{}{}.
\PrintBackRefs{\CurrentBib}

\bibitem [\protect \citeauthoryear {%
Sze%
, Chen%
, Yang%
\BCBL {}\ \BBA {} Emer%
}{%
Sze%
\ \protect \BOthers {.}}{%
{\protect \APACyear {2017}}%
}]{%
sze2017efficient}
\APACinsertmetastar {%
sze2017efficient}%
\begin{APACrefauthors}%
Sze, V.%
, Chen, Y\BHBI H.%
, Yang, T\BHBI J.%
\BCBL {}\ \BBA {} Emer, J\BPBI S.%
\end{APACrefauthors}%
\unskip\
\newblock
\APACrefYearMonthDay{2017}{}{}.
\newblock
{\BBOQ}\APACrefatitle {Efficient processing of deep neural networks: A tutorial and survey} {Efficient processing of deep neural networks: A tutorial and survey}.{\BBCQ}
\newblock
\APACjournalVolNumPages{Proceedings of the IEEE}{105}{12}{2295--2329}.
\PrintBackRefs{\CurrentBib}

\bibitem [\protect \citeauthoryear {%
Tiwari%
, Killamsetty%
, Iyer%
\BCBL {}\ \BBA {} Shenoy%
}{%
Tiwari%
\ \protect \BOthers {.}}{%
{\protect \APACyear {2022}}%
}]{%
tiwari2022gcr}
\APACinsertmetastar {%
tiwari2022gcr}%
\begin{APACrefauthors}%
Tiwari, R.%
, Killamsetty, K.%
, Iyer, R.%
\BCBL {}\ \BBA {} Shenoy, P.%
\end{APACrefauthors}%
\unskip\
\newblock
\APACrefYearMonthDay{2022}{}{}.
\newblock
{\BBOQ}\APACrefatitle {Gcr: Gradient coreset based replay buffer selection for continual learning} {Gcr: Gradient coreset based replay buffer selection for continual learning}.{\BBCQ}
\newblock
\BIn{} \APACrefbtitle {Proceedings of the IEEE/CVF Conference on Computer Vision and Pattern Recognition} {Proceedings of the ieee/cvf conference on computer vision and pattern recognition}\ (\BPGS\ 99--108).
\PrintBackRefs{\CurrentBib}

\bibitem [\protect \citeauthoryear {%
Van~de Ven%
\ \BBA {} Tolias%
}{%
Van~de Ven%
\ \BBA {} Tolias%
}{%
{\protect \APACyear {2019}}%
}]{%
van2019three}
\APACinsertmetastar {%
van2019three}%
\begin{APACrefauthors}%
Van~de Ven, G\BPBI M.%
\BCBT {}\ \BBA {} Tolias, A\BPBI S.%
\end{APACrefauthors}%
\unskip\
\newblock
\APACrefYearMonthDay{2019}{}{}.
\newblock
{\BBOQ}\APACrefatitle {Three scenarios for continual learning} {Three scenarios for continual learning}.{\BBCQ}
\newblock
\APACjournalVolNumPages{arXiv preprint arXiv:1904.07734}{}{}{}.
\PrintBackRefs{\CurrentBib}

\bibitem [\protect \citeauthoryear {%
van~de Ven%
, Tuytelaars%
\BCBL {}\ \BBA {} Tolias%
}{%
van~de Ven%
\ \protect \BOthers {.}}{%
{\protect \APACyear {2022}}%
}]{%
vandeven2022three}
\APACinsertmetastar {%
vandeven2022three}%
\begin{APACrefauthors}%
van~de Ven, G\BPBI M.%
, Tuytelaars, T.%
\BCBL {}\ \BBA {} Tolias, A\BPBI S.%
\end{APACrefauthors}%
\unskip\
\newblock
\APACrefYearMonthDay{2022}{}{}.
\newblock
{\BBOQ}\APACrefatitle {Three types of incremental learning} {Three types of incremental learning}.{\BBCQ}
\newblock
\APACjournalVolNumPages{Nature Machine Intelligence}{4}{}{1185--1197}.
\PrintBackRefs{\CurrentBib}

\bibitem [\protect \citeauthoryear {%
Wang%
, Zhang%
, Su%
\BCBL {}\ \BBA {} Zhu%
}{%
Wang%
\ \protect \BOthers {.}}{%
{\protect \APACyear {2023}}%
}]{%
wang2023comprehensive}
\APACinsertmetastar {%
wang2023comprehensive}%
\begin{APACrefauthors}%
Wang, L.%
, Zhang, X.%
, Su, H.%
\BCBL {}\ \BBA {} Zhu, J.%
\end{APACrefauthors}%
\unskip\
\newblock
\APACrefYearMonthDay{2023}{}{}.
\newblock
{\BBOQ}\APACrefatitle {A comprehensive survey of continual learning: Theory, method and application} {A comprehensive survey of continual learning: Theory, method and application}.{\BBCQ}
\newblock
\APACjournalVolNumPages{arXiv preprint arXiv:2302.00487}{}{}{}.
\PrintBackRefs{\CurrentBib}

\bibitem [\protect \citeauthoryear {%
Yosinski%
, Clune%
, Nguyen%
, Fuchs%
\BCBL {}\ \BBA {} Lipson%
}{%
Yosinski%
\ \protect \BOthers {.}}{%
{\protect \APACyear {2015}}%
}]{%
Yosinski_Clune_Nguyen_Fuchs_Lipson_2015}
\APACinsertmetastar {%
Yosinski_Clune_Nguyen_Fuchs_Lipson_2015}%
\begin{APACrefauthors}%
Yosinski, J.%
, Clune, J.%
, Nguyen, A.%
, Fuchs, T.%
\BCBL {}\ \BBA {} Lipson, H.%
\end{APACrefauthors}%
\unskip\
\newblock
\APACrefYearMonthDay{2015}{Jun}{}.
\newblock
{\BBOQ}\APACrefatitle {Understanding Neural Networks Through Deep Visualization} {Understanding neural networks through deep visualization}.{\BBCQ}
\newblock
\APACjournalVolNumPages{arXiv: Computer Vision and Pattern Recognition,arXiv: Computer Vision and Pattern Recognition}{}{}{}.
\PrintBackRefs{\CurrentBib}

\bibitem [\protect \citeauthoryear {%
Zeiler%
\ \BBA {} Fergus%
}{%
Zeiler%
\ \BBA {} Fergus%
}{%
{\protect \APACyear {2014}}%
}]{%
zeiler2014visualizing}
\APACinsertmetastar {%
zeiler2014visualizing}%
\begin{APACrefauthors}%
Zeiler, M\BPBI D.%
\BCBT {}\ \BBA {} Fergus, R.%
\end{APACrefauthors}%
\unskip\
\newblock
\APACrefYearMonthDay{2014}{}{}.
\newblock
{\BBOQ}\APACrefatitle {Visualizing and understanding convolutional networks} {Visualizing and understanding convolutional networks}.{\BBCQ}
\newblock
\BIn{} \APACrefbtitle {Computer Vision--ECCV 2014: 13th European Conference, Zurich, Switzerland, September 6-12, 2014, Proceedings, Part I 13} {Computer vision--eccv 2014: 13th european conference, zurich, switzerland, september 6-12, 2014, proceedings, part i 13}\ (\BPGS\ 818--833).
\PrintBackRefs{\CurrentBib}

\bibitem [\protect \citeauthoryear {%
Zenke%
, Poole%
\BCBL {}\ \BBA {} Ganguli%
}{%
Zenke%
\ \protect \BOthers {.}}{%
{\protect \APACyear {2017}}%
}]{%
zenke2017continual}
\APACinsertmetastar {%
zenke2017continual}%
\begin{APACrefauthors}%
Zenke, F.%
, Poole, B.%
\BCBL {}\ \BBA {} Ganguli, S.%
\end{APACrefauthors}%
\unskip\
\newblock
\APACrefYearMonthDay{2017}{}{}.
\newblock
{\BBOQ}\APACrefatitle {Continual learning through synaptic intelligence} {Continual learning through synaptic intelligence}.{\BBCQ}
\newblock
\BIn{} \APACrefbtitle {International conference on machine learning} {International conference on machine learning}\ (\BPGS\ 3987--3995).
\PrintBackRefs{\CurrentBib}

\end{thebibliography}

\end{document}